# Towards Explanation of DNN-based Prediction with Guided Feature Inversion

Mengnan Du, Ninghao Liu, Qingquan Song, Xia Hu
Department of Computer Science and Engineering, Texas A&M University
{dumengnan,nhliu43,song_3134,xiahu}@tamu.edu

## ABSTRACT

While deep neural networks (DNN) have become an effective computational tool, the prediction results are often criticized by the lack of interpretability, which is essential in many real-world applications such as health informatics. Existing attempts based on local interpretations aim to identify relevant features contributing the most to the prediction of DNN by monitoring the neighborhood of a given input. They usually simply ignore the intermediate layers of the DNN that might contain rich information for interpretation. To bridge the gap, in this paper, we propose to investigate a guided feature inversion framework for taking advantage of the deep architectures towards effective interpretation. The proposed framework not only determines the contribution of each feature in the input but also provides insights into the decision-making process of DNN models. By further interacting with the neuron of the target category at the output layer of the DNN, we enforce the interpretation result to be class-discriminative. We apply the proposed interpretation model to different CNN architectures to provide explanations for image data and conduct extensive experiments on ImageNet and PASCAL VOC07 datasets. The interpretation results demonstrate the effectiveness of our proposed framework in providing class-discriminative interpretation for DNN-based prediction.

## 1 INTRODUCTION

Deep neural networks (DNN) have achieved extremely high prediction accuracy in a wide range of fields such as computer vision [16, 21, 37], natural language processing [43], and recommender systems [17]. Despite the superior performance, DNN models are often regarded as black-boxes, since these models cannot provide meaningful explanations on how a certain prediction (decision) is made. Without the explanations to enhance the transparency of DNN models, it would become difficult to build up trust and credibility among end-users. Moreover, instead of accurately extracting insightful knowledge, DNN model may sometimes learn biases from the training data. Interpretability can be utilized as an effective debugging tool to find out the bias and vulnerabilities of models, analyze why it may fail in some cases, and ultimately figure out possible solutions to refine model performance.

Existing interpretation methods usually focus on two types of interpretations, i.e., model-level interpretation and instance-level interpretation. Model-level interpretation targets to find a good *prototype* in the input domain that is interpretable and can represent the abstract concept learned by a neuron or a group of neurons in a DNN [28]. Instance-level interpretation, however, tries to answer what features of an input lead it to activate the DNN neurons to make a specific prediction. As model-level interpretations usually delve into inherent model properties via in-depth theoretical analysis, instance-level interpretations are comparably more forthright, thus explain profound theories with more intuitive terms.

Conventional instance-level interpretations usually follow the philosophy of *local interpretation* [31]. Let $\mathbf{x}$ be an input for a DNN, the prediction of the DNN is denoted as a function $f(\mathbf{x})$. Through monitoring the prediction response provided by the function $f$ around the neighborhood of a given point $\mathbf{x}$, the features in $\mathbf{x}$ which cause a larger change of $f$ will be treated as more relevant to the final prediction. This either can be achieved by perturbing the input and observing the prediction differences [6, 11, 31, 45] (bottom-up manner) or calculating the gradient of output $f$ with respect to input $\mathbf{x}$ [36, 38–40] (top-down manner). Although these approaches locally interpret DNN predictions to some extent, they usually ignore the intermediate layers of DNN, thus leave out vast informative intermediate information [44, 47]. In addition, these methods have the risk of triggering the artifacts of DNN models [15, 22]. It has been demonstrated that some generated inputs can fool DNN and lead DNN to make unexpected outputs, which can not be counted as meaningful interpretations. By taking advantage of the intermediate layers information, it is more likely to characterize the behaviors of DNN under normal operating conditions. It motivates us to explore the utilization of intermediate information to derive more accurate interpretations.

Feature inversion has been initially studied for visualizing and understanding intermediate feature representations of DNN [8, 27]. It has been shown that the CNN representation could be inverted to an image which sheds light on the information extracted by each convolutional layer. The inversion results indicate that as the information propagates from the input layer to the output layer, the DNN classifier gradually compresses the input information, and discard information irrelevant to the prediction task. Besides, the inversion result from a specific layer also reveals the amount of information contained in that layer. However, these inversion results are relatively rough and obscure for delicate interpretations. It remains challenging to automatically extract the contributing factors for prediction, i.e., the location for the target object, in the input utilizing the feature inversion.

In this paper, we propose an instance-level DNN interpretation model by performing guided image feature inversion. Leveraging the observations found in our preliminary experiments that the higher layers of DNN do capture the high-level content of the input as well as its spatial arrangement, we present guided feature reconstructions to explicitly preserve the object localization information in a "mask", so as to provide insights of what information is actually employed by the DNN for the prediction. In order to induce class-discriminative power upon interpretations, we further establish connections between the input and the target object by fine-tuning the interpretation result obtained from guided feature inversions with class-dependent constraints. In addition, we show that the intermediate activation values at higher convolutional layers of DNN are able to behave as a stronger regularizer, leading to more smooth, and continuous saliency maps. This regularization dramatically decreases the possibility to produce artifacts, thus providing more exquisite interpretations. Four major contributions of this paper are summarized as follows:

- We propose a novel guided feature inversion method to provide instance-level interpretations of DNN. The proposed method could locate the salient foreground part, thus determining which part of information in the input instance is preserved by the DNN, and which part is discarded.
- Through adding class-dependent constraints upon the guided feature inversion, the proposed method could provide class-discriminative power for more exquisite interpretations.
- Leveraging the integration of the intermediate activation values as masks, we further lower the possibility to produce artifacts and increase the optimization efficiency.
- Experimental results on three image datasets validate that the proposed method could accurately localize discriminative image regions which is consistent with human cognition.

The rest of this paper is organized as follows. Section 2 summarizes two lines of work related to this paper. Section 3 introduces the proposed framework for interpreting DNN-based predictions. Section 4 presents experimental results to verify the effectiveness of our proposed framework. Section 5 gives the conclusional remarks.

## 2 RELATED WORK

There are a number of techniques developed to interpret machine learning models [7, 12, 23, 24, 31]. Among them, two lines of work are most relevant to this paper: visualizing DNN feature representation through feature inversion, and instance-level interpretation for DNN-based prediction. We present brief reviews of these two research fields as follows.

**Feature inversion** Mahendran and Vedaldi [27] inverted intermediate CNN feature representation [4] from different layers in order to have some insights into the working mechanism of CNN. The up-convolutional neural network was utilized in [8] to invert the intermediate features of CNN and led to more accurate image reconstruction, revealing that rich information is contained in these inner features. Feature inversion has become an effective tool to be applied to high-level image generation and transformation tasks, due to its power to generate high-quality visualizations. For instance, feature inversion was utilized to separate the content and style of any natural images, and a new image was then synthesized with content and style derived from two source images [13]. Upchurch *et al.* [42] first linearly interpolated the higher layer of CNN and then inverted the modified feature to an image to perform high-level semantic transformations.

**Interpretation for DNN-based prediction** A large fraction of existing interpretation methods are based on sensitivity analysis, i.e., calculating the sensitivity of the classification output in terms of the input. A significant prediction probability drop with a certain class means a big feature importance towards the prediction. This type of methods can be further classified into two categories: gradient based methods, and perturbation based methods.

Gradient based methods computed the partial derivative of the class score with respect to the input image using backpropagation [36]. Integrated gradient [40] estimated the global importance of each pixel to the prediction, instead of the local sensitivity. Guided back-prorogation [39] modified the gradient of RELU (rectified linear units) function by discarding negative values at the backpropagation process. Smooth Grad [38] addressed the visual noise problem of gradient based interpretation method by introducing noise to the input. Gradient based methods are advantageous in that they are computationally efficient, *i.e.*, using few forward and backward iterations is sufficient to generate an interpretation saliency map. However, the saliency maps are typically blurry and may mistakenly highlight the background which is irrelevant to the target object, as can be seen from the visualization on Fig. 2.

The philosophy of perturbation based interpretation is perturbing the original input and observing the prediction probability of the DNN model. Through measuring the prediction difference, the attributing factors in the input with the class label can thus be located. The image patches [45] were used to occlude the input image in the form of sliding window, the relevance of each patch was then calculated through the drop of CNN prediction probability. Similarly, super-pixel occlusion can also be utilized in the LIME [31] model, which learned a local linear model to obtain the contributing score of each superpixel. The recent work of [11] and [6] utilized a mask to perturb the input image, and learn the mask using gradient descent. Perturbation based approaches yield more visually pleasing explanations comparing to the results generated by gradient based methods. Still, these approaches are highly vulnerable to surprising artifacts. The perturbation may produce inputs that are totally different from the natural image statistics in the training set. The predictions for such inputs thus skew a lot which eventually lead to uninterpretable explanations.

Besides these two types of interpretation, some recent work proposed to provide explanations through investigation on hidden layers of DNN. Both CAM [48] and Grad-CAM [34] generated interpretation saliency maps by combining the feature maps in the intermediate layers. The difference is that the former can only be applied to a small subset of CNN classifiers that have global average pooling layer prior to the output layer, while the latter integrates intermediate features using gradient, and thus can be applied to a wider range of CNN architectures.

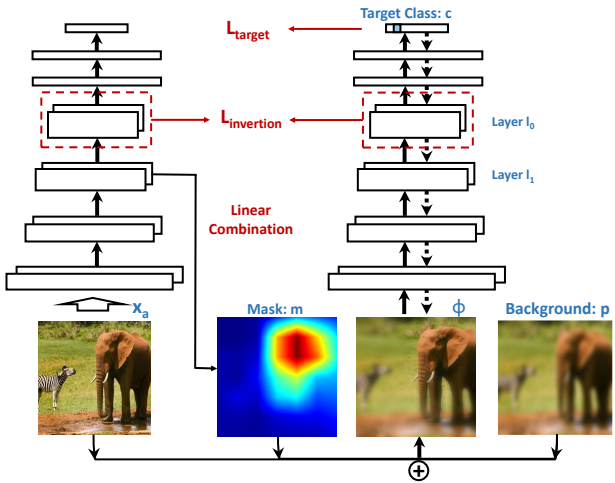

Figure 1: An illustration of the proposed interpretation framework. First, the original input $\mathbf{x}_a$ is sent to the CNN (on the left), and the representation at each layer of CNN is calculated and saved. Second, class-discriminative interpretation result is obtained by interacting with the CNN (on the right). The guided feature inversion $\Phi$ extracts the location for all the foreground objects (Sec. 3.2). Then we fine-tune the inversion result using the activation of the neuron for the target class in the last layer of DNN (Sec. 3.3). Besides, we impose a strong regularizer by using the integration of the intermediate layer activations of the original input as the mask (Sec. 3.4).

Although our method shares some similarities with the work that incorporate intermediate network activations into interpretation, *i.e.*, CAM [48] and Grad-CAM [34], they are significantly different. These two methods combine the channels of intermediate layers heuristically. In contrast, we optimize the combination of these channels to make the reconstructed input having consistent intermediate feature representation with the original input. It makes our approach yielding interpretable explanations which truly reflect the the decision making process of DNN. Furthermore, by taking advantage of the regularization power of guided feature inversion, our method dramatically decreases the possibility to produce artifacts, thus providing more exquisite interpretations.

## 3 INTERPRETATION OF DNN-BASED PREDICTION

In this section, we introduce the proposed interpretation framework for interpreting DNN-based predictions. The pipeline of the proposed framework is illustrated in Fig. 1. The main idea of the proposed framework is to identify the image regions that simultaneously encode the location information of the target object and match the feature representation of the original image. Moreover, we focus on class-dependent interpretation by formulating a constraint to force the interpretation result to strongly activate the neuron corresponding to the target class at the last layer of DNN. Besides, the image regions are captured by a mask obtained by integrating the intermediate activation values at higher convolutional layers of the DNN, in order to reduce undesirable artifacts and ensure that the interpretation results are meaningful. The details of the proposed framework are discussed as below.

### 3.1 Problem Statement

We first introduce the basic notations used in this paper. Considering a multiclass classification task, a pre-trained DNN model can be treated as a function $\mathbf{f}(\mathbf{x})$ of the input $\mathbf{x} \in \mathbb{R}^d$. When feeding an input $\mathbf{x}$ to the DNN model, $\mathbf{f}_c(\mathbf{x}) \in [0, 1], c \in \{1, ..., C\}$ represents the corresponding classification probability score for class $c$. We focus on post-hoc interpretations [28] through explaining the DNN prediction result for a given data instance $\mathbf{x}$ to ensure the generality of the proposed method. We aim to find out the contributing factors in the input $\mathbf{x}$ that lead the DNN to make the prediction. Specifically, let $c$ be the target object class that we want to interpret, and $\mathbf{x}_i$ corresponds to the $i^{th}$ feature, then the interpretation for $\mathbf{x}$ is encoded by a score vector $\mathbf{s} \in \mathbb{R}^d$ where each score element $\mathbf{s}_i \in [0, 1]$ represents how relevant of that feature is for explaining $\mathbf{f}_c(\mathbf{x})$. We use image classification as an example in this paper. In this case, the input vector $\mathbf{x}_a$ corresponds to the pixels of an image, and the score vector $\mathbf{s}$ will be the saliency map (or attribution map), where the pixels with higher scores represent higher relevance for the classification task.

### 3.2 Interpretation through Feature Inversion

In this section, we derive the initial interpretation for DNN-based interpretation using guided feature inversion. It has been studied that the deep image representation extracted from a layer in CNN could be inverted to a reconstructed image which captures the property and invariance encoded in that layer [8, 27]. An observation is that feature inversion can reveal how much information is preserved in the feature at a specific layer. Specifically, the reconstruction results from features of the first few layers preserve almost all the detailed image information, while the inversions from the last few layers merely contain the rough shape of the original image. This observation shows that CNN gradually filters out unrelated information for the classification task as the layer goes deeper. It thus motivates us to explore the feature inversion of higher layers of CNN to provide explanation for classification result of each instance.

Given a pre-trained $L$-layers CNN model, the intermediate feature representation at layer $l \in \{1, 2, ..., L\}$ could be denoted as a function $\mathbf{f}^l(\mathbf{x}_a)$ of the input image $\mathbf{x}_a$. The process of inverting the feature representation at layer $l_0$ can be regarded as computing the approximated inversion $\mathbf{f}^{-1}$ of the representation $\mathbf{f}^{l_0}(\mathbf{x}_a)$. The feature inversion tries to find the image $\mathbf{x}$ that minimizes the following objective function:

$$\mathbf{x}^* = \underset{\mathbf{x}}{\operatorname{argmin}} \|\mathbf{f}^{l_0}(\mathbf{x}) - \mathbf{f}^{l_0}(\mathbf{x}_a)\|^2 + \mathcal{R}(\mathbf{x}), \quad (1)$$

where the squared error term forces the representation $\mathbf{f}^{l_0}(\mathbf{x}^*)$ of inversion result $\mathbf{x}^*$ and the original input representation $\mathbf{f}^{l_0}(\mathbf{x}_a)$ to be as similar as possible, while the regularization term $\mathcal{R}$ imposes a natural image prior. As the layer goes deeper for inversion, it is with higher confidence about which part of the input information is ultimately preserved for final prediction. Since the spatial configuration information of the target object is discarded at *fully connected layers*, the feature inversion cannot recover the accurate object localization information from these layers [8, 27]. Therefore, we fix the inversion layer $l_0$ to be the last pooling layer before the first fully connected layer. Take the 8-layer AlexNet [21] as an example. We use the *pool5 layer* for feature inversion, where the spatial size is (6 × 6) and the total number of channels of this layer is 256. $\mathbf{x}$ is initialized randomly, and the optimal inversion result $\mathbf{x}^*$ could be obtained using gradient descent.

The contribution scores for quantifying the correlation between input pixels and outputs can be determined from the inversion result $\mathbf{x}^*$. A straightforward way to find out the contributing factors in the input is to calculate the pixel-wise difference between $\mathbf{x}_a$ and $\mathbf{x}^*$. The resulting saliency map $\mathbf{s}$ can thus be computed as: $\mathbf{s} = \frac{\mathbf{x}_a - \mathbf{x}^*}{\mathbf{x}_a}$. However, this is not feasible in practice due to the reason that the saliency map is noisy, where even adjacent pixels with similar color and texture patterns could have distinct saliency scores. Thus, this formulation may not truly represent the contributions of each pixel. To tackle this problem, we propose the *guided feature inversion method*, where the expected inversion image representation is reformulated as the weighted sum of the original image $\mathbf{x}_a$ and another noise background image $\mathbf{p}$. We replace the optimization target $\mathbf{x}$ in Eq. (1) with the guided inversion $\Phi(\mathbf{x}_a, \mathbf{m})$, which is formulated as follows:

$$\Phi(\mathbf{x}_a, \mathbf{m}) = \mathbf{x}_a \odot \mathbf{m} + \mathbf{p} \odot (1 - \mathbf{m}). \quad (2)$$

A desirable property of the above formula is that the inversion image representation $\Phi(\mathbf{x}_a)$ is located in the natural image space manifold. To this end, we consider three choices of $\mathbf{p}$: a grayscale image in which the value of each pixel is set to the average color over ImageNet dataset (the normalized values for the three channels of RGB color space are [0.485, 0.456, 0.406]), a Gaussian white noise image, or a blurred image which is obtained by applying a Gaussian blur filter to the original image $\mathbf{x}_a$ [11]. The weight vector $\mathbf{m} \in [0, 1]^d$ denotes the significance of each pixel contributing to the feature representation $\mathbf{f}^l(\mathbf{x}_a)$. Therefore, we can recover the object location information from the weight vector $\mathbf{m}$. Instead of directly finding the inversion image representation, we optimize the weight vector $\mathbf{m}$, which is formulated as follows:

$$L_{\text{inversion}}(\mathbf{x}_a, \mathbf{m}) = \|\mathbf{f}^{l_0}(\Phi(\mathbf{x}_a, \mathbf{m})) - \mathbf{f}^{l_0}(\mathbf{x}_a)\|^2 + \alpha \cdot \frac{1}{d} \sum_{i=1}^{d} \mathbf{m}_i. \quad (3)$$

The first term corresponds to the inversion error. The error will be zero if all entries within $\mathbf{m}$ equal to 1. In the second term, we limit the area of $\mathbf{m}$ to be as small as possible in order to find out the most contributing regions in input $\mathbf{x}_a$. The parameter $\alpha$ is utilized to balance the inversion error and the area of $\mathbf{m}$. This formulation not only creates image $\Phi(\mathbf{x}_a, m)$ which matches the inner feature representation at layer $l_0$, but also preserves the object localization information in $\mathbf{m}$.

### 3.3 Class-Discriminative Interpretation

In this section, we derive class-discriminative interpretation to distinguish different categories of objects from the generated mask $\mathbf{m}$. Up to now, we only use the information from the former convolutional layers of CNN to generate mask $\mathbf{m}$. Although it has extracted all the foreground object information which are crucial for subsequent prediction, the connection between the input $\mathbf{x}_a$ and the target label $c$, which is largely encoded in the rest layers, has not been established yet. On the other hand, one image may contain multiple foreground objects. Lacking discriminative power, the mask $\mathbf{m}$ derived from the aforementioned formulation Eq.(3) will highlight all these objects. For instance, the mask $\mathbf{m}$ in Fig. 3 (b) extracts both the locations of zebra and elephant, rather than provides localization for only the object class $c$.

We would like the interpretation result to highlight the target class and suppress the irrelevant classes through further leveraging the non-utilized layers. To this end, we render the guided feature reconstruction result $\Phi(\mathbf{x}_a)$ to strongly activate the softmax probability value $\mathbf{f}_c^L$ at the last hidden layer $L$ of CNN for a given target label $c$, and reduce the activation for other classes $\{1, ..., C\} \setminus c$. In the meantime, we formulate the complementary counterpart of the mask $\mathbf{m}$ as $\mathbf{m}_{bg} = 1 - \mathbf{m}$. It contains irrelevant information with respect to target class $c$, including image background and other classes of foreground objects. Using the heatmap $\mathbf{m}_{bg}$ as weight, the background part of the image can be calculated as the weighted sum of the original images $\mathbf{x}_a$ and $\mathbf{p}$:

$$\Phi_{bg}(\mathbf{x}_a, \mathbf{m}_{bg}) = \mathbf{x}_a \odot \mathbf{m}_{bg} + \mathbf{p} \odot (1 - \mathbf{m}_{bg}). \quad (4)$$

We expect the object information for the target class $c$ to be removed from $\Phi_{bg}(\mathbf{x}_a, \mathbf{m})$ to the maximization degree. As such, when feeding $\Phi_{bg}(\mathbf{x}_a, \mathbf{m})$ to the CNN classifier, the prediction probability is supposed to be small. The class-discriminative interpretation formulation is defined as follows:

$$L_{\text{target}}(\mathbf{x}_a, \mathbf{m}) = -\mathbf{f}_c^L(\Phi(\mathbf{x}_a, \mathbf{m})) + \lambda \mathbf{f}_c^L(\Phi_{bg}(\mathbf{x}_a, \mathbf{m})) + \beta \cdot \frac{1}{d} \sum_{i=1}^{d} \mathbf{m}_i, \quad (5)$$

where $\lambda$ and $\beta$ control the importance of the highlighting term, suppressing term and the area of $\mathbf{m}$.

### 3.4 Regularization by Utilizing Intermediate Layers

The aforementioned formulation still has the weakness of generating undesirable artifacts without regularizations imposed to the optimization process. Lee and Verleysen pointed out that image data lie on a low-dimensional manifold [22]. However, it is possible that the generated $\mathbf{m}$ could push $\Phi(\mathbf{x}_a, \mathbf{m})$ out of valid data manifold, where the DNN classifier does not work properly [10]. The generated $\mathbf{m}$ could be composed of some disconnected and noisy patches, and no patterns about the target object could be identified from it. To generate more meaningful interpretation, we propose to exquisitely design the regularization term of the mask $\mathbf{m}$ to overcome the artifacts problem. Existing works have shown that the visualization performance of the interpretation results could be improved by inducing $\alpha$-norm [27], total variation norm [27], and Gaussian blur [44]. Though these regularizations could reduce the

**Algorithm 1:** Interpretation through guided feature inversion.

**Input:** $\mathbf{x}_0$, CNN model $f$.
**Output:** $\mathbf{m}, \omega$.

1. Initialize the parameter $\omega_i = 0.1, i \in \{1, ..., n\}, \gamma = 10$, iteration numbers $max\_iter = 10, \eta = 10^{-2}, t = 0$;
2. $\mathbf{p}$ is obtained by applying Gaussian blurred of radius 11 to $\mathbf{x}_0$;
3. Set the inversion layer $l_0$, and set the base channel layer $l_1$;
4. **while** $t \leq max\_iter$ **do**
5.     $\mathbf{m}_t = \sum_i \omega_{t,i} \cdot \mathbf{f}_i^{l_1}(\mathbf{x}_a)$;
6.     $\mathbf{m}_t \leftarrow \text{Upsample}(\frac{\mathbf{m}_t - \min(\mathbf{m}_t)}{\max(\mathbf{m}_t) - \min(\mathbf{m}_t)})$;
7.     $\Phi(\mathbf{x}_a, \mathbf{m}_t) = \mathbf{x}_a \odot \mathbf{m}_t + \mathbf{p} \odot (1 - \mathbf{m}_t)$;
8.     $L_{\text{inversion}} = \|\mathbf{f}^{l_0}(\Phi(\mathbf{x}_a, \omega_t)) - \mathbf{f}^{l_0}(\mathbf{x}_a)\|^2 + \gamma \cdot \|\omega_t\|_1$;
9.     $\omega_{t+1} = Adam(L_{\text{inversion}}, \eta)$;
10.     $\omega_{t+1} \leftarrow \text{Clip}(\omega_{t+1}, 0, \infty)$;
11.     $t = t + 1$;
12. $\mathbf{m} = \sum_i \omega_{t,i} \mathbf{f}_i^l(\mathbf{x}_a)$;
13. **return** $\mathbf{m}, \omega$.

**Algorithm 2:** Class-discriminative interpretation.

**Input:** $\mathbf{x}_0$, CNN model $f$, target label $c$.
**Output:** $\mathbf{m}, \omega$.

1. Initialize the parameter $\omega$ to be the result returned in Algorithm 1, $\lambda = 1, \delta = 1$, iteration numbers $max\_iter = 70, \eta = 10^{-2}, t = 0$;
2. $\mathbf{p}$ is obtained by applying Gaussian blurred of radius 11 to $\mathbf{x}_0$;
3. Set the base channel layer $l_1$;
4. **while** $t \leq max\_iter$ **do**
5.     $\mathbf{m}_t = \sum_i \omega_{t,i} \cdot \mathbf{f}_i^{l_1}(\mathbf{x}_a)$;
6.     $\mathbf{m}_t \leftarrow \text{Upsample}(\frac{\mathbf{m}_t - \min(\mathbf{m}_t)}{\max(\mathbf{m}_t) - \min(\mathbf{m}_t)})$;
7.     $\Phi(\mathbf{x}_a, \mathbf{m}_t) = \mathbf{x}_a \odot \mathbf{m}_t + \mathbf{p} \odot (1 - \mathbf{m}_t)$;
8.     $\Phi_{bg}(\mathbf{x}_a, \mathbf{m}_t) = \mathbf{x}_a \odot (1 - \mathbf{m}_t) + \mathbf{p} \odot \mathbf{m}_t$;
9.     $L_{\text{target}} = -\mathbf{f}_c^L(\Phi(\mathbf{x}_a, \omega)) + \lambda \mathbf{f}_c^L(\Phi_{bg}(\mathbf{x}_a, \omega)) + \delta \cdot \|\omega\|_1$;
10.     $\omega_{t+1} = Adam(L_{\text{target}}, \eta)$;
11.     $\omega_{t+1} \leftarrow \text{Clip}(\omega_{t+1}, 0, \infty)$;
12.     $t = t + 1$;
13.     $\eta \leftarrow \frac{\eta}{2}$ if $t \% 10 = 0$;
14. $\mathbf{m} = \sum_i \omega_{t,i} \mathbf{f}_i^l(\mathbf{x}_a)$;
15. **return** $\mathbf{m}, \omega$.

occurrence of unwanted artifacts to some extent, only limited performances are achieved according to our preliminary experiments since they simply smooth the interpretation results.

To address this problem, we impose a stronger natural image prior by utilizing the intermediate activation features of CNN. This is motivated by the fact that higher convolutional layers of CNN are responsive to specific and semantically meaningful natural part (*e.g.*, face, building, or lamp) [44, 47]. For instance, as evidenced by [44], the ($13 \times 13$) activations for the $151^{st}$ channel of *conv5* for a CNN responds to animal and human faces. More importantly, the activations on these layers preserve the information about object locations. These feature layers, when projected down to the pixel space, could correspond to the rough location of these semantic parts. Similar to decomposing an object into the combinations of its high-level parts, we assume the mask $\mathbf{m}$ could be decomposed as the combination of channels at a high-level layer of the targeted CNN model. Specially, let $\mathbf{f}_i^{l_1}(\mathbf{x}_a)$ represent the $i^{th}$ channel of the $l_1^{th}$ layer of the CNN. We build the weight mask $\mathbf{m}$ as the weighted sum of the channels at a specific layer $l_1$:

$$\mathbf{m} = \sum_i \omega_i \mathbf{f}_i^{l_1}(\mathbf{x}_a). \tag{6}$$

Parameter vector $\omega$ captures the relevance of each channel map for the final prediction. Since the activation values $\mathbf{f}^{l_1}(\mathbf{x}_a)$ could locate at various ranges, the generated mask $\mathbf{m}$ may take a wide range of values if no constraint is imposed to the parameter $\omega$. It is essential to limit $\mathbf{m} \in [0, 1]^n$, so as to guarantee the guided reconstruction in Eq.(2) is constrained at the expected input domain range. Therefore, the mask is further normalized through Min-Max normalization:

$$\mathbf{m} \leftarrow \frac{\mathbf{m} - \min(\mathbf{m})}{\max(\mathbf{m}) - \min(\mathbf{m})}. \tag{7}$$

Before applying to the objective function, we still need to enlarge the mask $\mathbf{m}$ from the small resolution to an identical resolution with the original input. To guarantee the smoothness of the enlarged representation $\mathbf{m}$, we apply upsampling using bilinear interpolation. After replacing the original mask at Eq.(2) and Eq.(4) with the newly derived mask, we expect the guided inversion representation $\Phi(\mathbf{x}_a, \mathbf{m})$ and the background representation $\Phi_{bg}(\mathbf{x}_a, m)$ to become less likely to be affected by artifacts. Finally, the interpretation objective Eq.(3) and Eq.(5) could be reformulated as:

$$L_{\text{inversion}}(\mathbf{x}_a, \omega) = \|\mathbf{f}^{l_0}(\Phi(\mathbf{x}_a, \omega)) - \mathbf{f}^{l_0}(\mathbf{x}_a)\|^2 + \gamma \cdot \|\omega\|_1, \tag{8}$$

$$L_{\text{target}}(\mathbf{x}_a, \omega) = -\mathbf{f}_c^L(\Phi(\mathbf{x}_a, \omega)) + \lambda \mathbf{f}_c^L(\Phi_{bg}(\mathbf{x}_a, \omega)) + \delta \cdot \|\omega\|_1, \tag{9}$$

where $\omega_i \geq 0, i \in \{1, ..., n\}$, since we only focus on the channel maps which have a positive influence for making a prediction. Parameters $\gamma$ and $\delta$ control the importance of the regularization term. We utilize the $\ell_1$-norm regularization to ensure that only very few entries in the parameter vector $\omega$ is non-zero. This is motivated from the observation that objects could be depicted using only one or few object parts. The $\ell_1$-norm regularization also drastically reduces the possibility of model over-fitting. This natural image prior brings two benefits. On one hand, it guarantees that less artifacts will be produced in the optimized mask. On the other hand, it dramatically reduce the numbers of the parameters to be optimized, leading to increased efficiency of the optimization.

We apply a two-stage optimization to derive class-discriminative interpretation for DNN-based prediction. In the first stage, we perform interpretation using guided feature inversion to find out the salient foreground part (see Algorithm 1). After initializing the parameter $\omega_i = 0.1, i \in \{1, ..., n\}$, we perform gradient descent optimization to find out the optimal parameter vector $\omega$ as well as the mask $\mathbf{m}$. In the second stage, we obtain the class-discriminative interpretation by further fine-tune the parameter vector $\omega$ (see Algorithm 2). After initializing $\omega$ to be the result obtained in Algorithm 1, we reduce the learning rate every 10 iterations and further fine-tune the parameter $\omega$ in order to let the mask $\mathbf{m}$ more relevant to the target label. Note that to tackle the constraint in objective function (8) and (9) that parameter $\omega_i \geq 0, i \in \{1, ..., n\}$, we simply clip the $\omega$ to the valid range after each gradient descent iteration. At last, we generate the interpretation mask $\mathbf{m}$ for target class $c$.

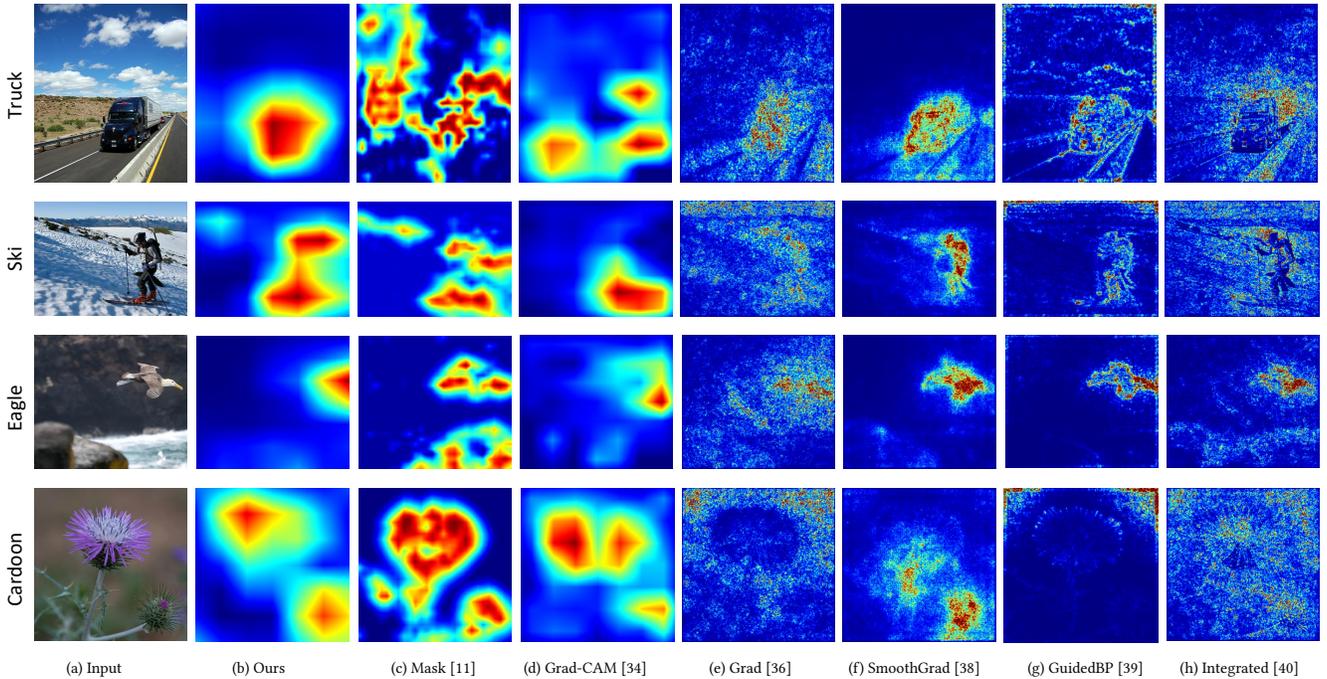

Figure 2: Visualization saliency maps comparing with 6 state-of-the-art methods.

## 4 EXPERIMENTS

In this section, we conduct experiments to evaluate the effectiveness of the proposed interpretation method. First, we visualize the interpretation results on ImageNet dataset in Sec. 4.1 under various settings of model architectures. Second, we test the localization performance by applying it to weakly supervised object localization task in Sec. 4.2. Third, we discuss the class discriminability of our algorithm in Sec. 4.3. Finally, we evaluate the guided feature inversion part of the proposed method by applying it to salient object detection task in Sec. 4.4.

### 4.1 Visualization of Interpretation Results

*4.1.1 Experimental Settings.* In the following experiments, unless stated otherwise, the interpretation results are provided based on VGG-19 [37]. Specifically, we utilize the pre-trained VGG-19 model from torchvision[1]. Its Top-1 prediction error and Top-5 prediction error on ImageNet dataset are 27.62% and 9.12%, respectively. For inversion layer $l_0$, we use the *pool5* layer (the 5th pooling layer), and the size of representation at this layer is $(7 \times 7 \times 512)$. As for the base channel layer $l_1$ in Eq. (6), we utilize *conv5_4*, which is the layer prior to *pool5*, and has 512 channels of size $(14 \times 14)$. Therefore the length of parameter vector $\omega$ is also 512. The entries of $\omega$ are all initialized as 0.1. The $\lambda$ is fixed to 1. The $\ell_1$-norm regularizer parameter $\gamma$ and $\delta$ are set to 10 and 1, respectively. All these parameters are tuned based on the quantitative and qualitative performance of the interpretation on a subset of the ILSVRC2014 [33] training set.

For input images, we resize them to the shape $(224 \times 224 \times 3)$, and transform them to the range $[0, 1]$, and then normalize them using mean vector $[0.485, 0.456, 0.406]$ and standard deviation vector $[0.229, 0.224, 0.225]$. No further pre-processing is performed. The background image **p** is obtained by applying a single Gaussian blur of radius 11 to the original input.

[1]http://pytorch.org/docs/master/torchvision/models.html

We employ Adam [19] optimizer to perform gradient descent, which achieves faster convergence rate than stochastic gradient descent (SGD). The number of iteration steps is 10, and 70 for the first stage and the second stage respectively. The learning rate is initialized to $10^{-2}$ for Adam optimizer, chosen by line search. At the second stage, we apply step decay, and reduce the learning rate by half every ten epochs.

*4.1.2 Visualization.* We qualitatively compare the saliency maps produced using the proposed method with those produced by six state-of-the-art methods, including Grad [36], GuidedBP [39], SmoothGrad [38], Integrated [40], Mask [11], Grad-CAM [34], see Fig. 2. Comparing to the other methods, it shows that our method generates more visually interpretable saliency maps. Take the second row for example. The DNN predicts the *ski* category with 98.2% confidence, and our method accurately highlights the helmet, skis, as well as ski poles. At the fourth row, our method highlights both the dominant cardoon at the center and the smaller cardoon at the lower right corner.

We also demonstrate that the proposed interpretation method could distinguish different classes as shown in Fig. 3. The VGG-19 model classifies the input as *African elephant* with 95.3% confidence, and *zebra* with 0.2% confidence, our model correctly gives the interpretation locations for both of two labels, even though the prediction probability of the latter is much lower than the probability of the former. An interesting discovery obtained from the saliency maps is that the head, ear, and nose parts are most discriminative to distinguish elephant, while the body part is most crucial to classify zebra. It is consistent with our human cognition since we also rely on the head shape of the elephant and the black-and-white striped coats of zebra to classify them. Thus this interpretation is able to build trust with end users [5].

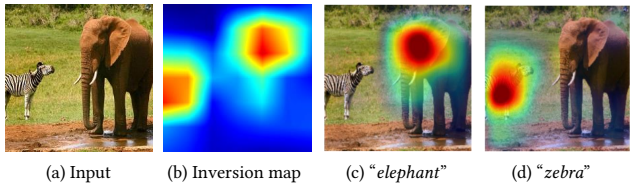

(a) Input    (b) Inversion map    (c) "*elephant*"    (d) "*zebra*"

**Figure 3: Class discriminability of our algorithm. The inversion result (b) highlights all the foreground objects, while the final interpretation (c) and (d) successfully highlight only the target object.**

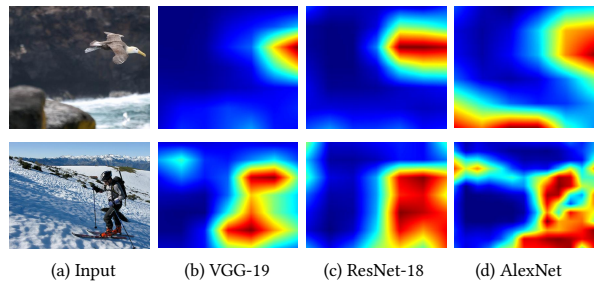

(a) Input    (b) VGG-19    (c) ResNet-18    (d) AlexNet

**Figure 5: Interpretation results for three DNN architectures.**

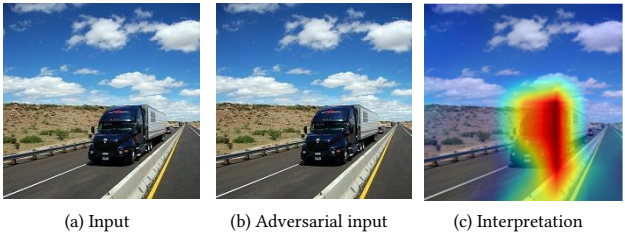

(a) Input    (b) Adversarial input    (c) Interpretation

**Figure 4: Interpretation result for an adversarial example. (a) Original input: *trailer truck* with 95.8% confidence, (b) Adversarial input: *container ship* with 48.0% confidence, (c) Interpretation of the adversarial input for the target class *trailer truck*.**

*4.1.3 Interpretation Results Under Adversarial Attacks.* We test whether our interpretation method could tolerate adversarial attacks [25]. Ghorbani *et al.* [14] demonstrated that several recent interpretation models, including [20, 35, 36, 40], are fragile under adversarial attack. A small perturbation in the input would drastically change their interpretation results. Specifically, Fast Gradient Sign method [15] is utilized to produce adversarial inputs. Fig. 4 illustrates the interpretation result for an adversarial example. After adding some small and unnoticeable perturbation to the original input, the adversarial attack [15] causes the classifier to miscategorize the input as *container ship* with high confidence (48.0%). Our interpretation model still could give the location for the true label *tailer truck*. It demonstrates that the proposed interpretation method is quite robust, and could provide reasonable interpretation under adversarial setting.

*4.1.4 Interpretation Results Under Different CNN Architectures.* Besides VGG-19 [37], we also provide explanation for two different network architectures, including AlexNet [21] and ResNet-18 [16]. For AlexNet, the *pool5* layer is utilized for both the inversion layer $l_0$ as well as the base channel layer $l_1$, the representation size of which is (6×6×256). As for ResNet-18, the inversion layer $l_0$ utilizes *pool5*, and the base channel layer $l_1$ utilizes the next to last *conv layer*, both have size of (7 × 7 × 512). For the rest of parameters, we utilize the same configuration as VGG-19 (see Sec.4.1.1).

The interpretation visualizations of three CNN architectures are shown in Fig. 5. For both the two inputs, the saliency maps generated by VGG-19 and ResNet-18 give the accurate location, while the ones yielded by AlexNet also highlight part of the background. One possible reason is that the AlexNet has only half number of channels at layer $l_1$ compared to the other two architectures. Besides, the smaller kernel filters as well as the increasing number of layers enable VGG-19 and ResNet-18 to learn more complex features and also lead to higher localization accuracy than AlexNet. It demonstrates that our model can be applied to a wide range of network architectures, including the neural network with skip-layer connections, and without fully connected layers.

In addition, we analyze the running efficiency of our interpretation method under the three CNN architectures. Processing an instance takes an average time of 8.04, 6.43, 4.82 seconds for VGG-19, ResNet-18, and AlexNet respectively, using GPU implementation of Pytorch, and with the batch size setting to 1. The running time is consistent with the model complexity. Since VGG-19 has more parameters than the other two CNN architectures, it spends longer computation time to provide explanation. In addition, increasing the batch size is supposed to dramatically improve the interpretation efficiency of our method.

## 4.2 Quantitative Evaluation via Weakly Supervised Object Localization

In this section, we evaluate the localization performance of our interpretation method by applying the generated saliency maps to weakly supervised object localization tasks. The experiments are performed on the ImageNet validation set, which contains 50,000 images with bounding box annotations. Similar to [11, 46] as well as ILSVRC2014 [33] setting, 1762 images are excluded from the evaluation task because of their pool quality of annotations.

The saliency maps are binarized using mean thresholding by $\alpha \cdot m_I$, where $m_I$ is the mean intensity of the saliency map and $\alpha \in [0.0 : 0.5 : 10.0]$, using the same setting with [11]. The tightest rectangle enclosing the whole segmented saliency map is counted as the final bounding box. The IOU (intersection over union) metric is utilized to measure the localization performance of each input instance. The localization is considered to be successful if the IOU score for an instance exceeds 0.5, otherwise it is treated as an error. The weakly supervised error is judged by the average localization error over the ImageNet validation set. For each comparing method, the $\alpha$ value is tuned using 1000 images selected from the ILSVRC2014 training dataset.

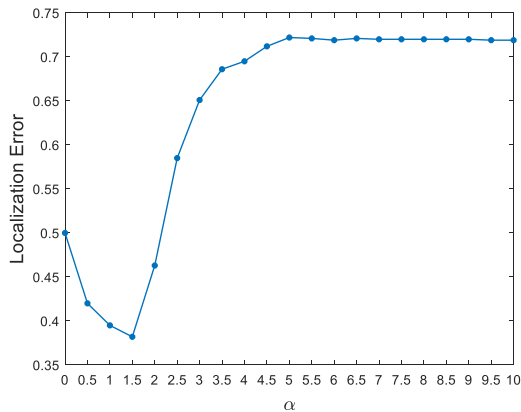

Figure 6: Localization error curve under different $\alpha$ values.

|   | Grad | GuidedBP | LRP | CAM | Mask | Real | Ours |
|---|---|---|---|---|---|---|---|
| $\alpha$ | 5.0 | 4.5 | 1.0 | 1.0 | 0.5 | - | 1.1 |
| Error(%) | 41.7 | 42.0 | 57.8 | 48.1 | 43.2 | 36.9 | 38.2 |

Table 1: Localization errors, and the optimal $\alpha$ values on ImageNet validation set of comparing methods. Error rate of comparing methods are taken from [11].

*4.2.1 Comparison with Other Methods.* The object localization performance of the proposed method is compared with those of six state-of-the-art methods, including Grad [36], GuidedBP [39], LRP [2], Mask [11], CAM [48], and Real-time saliency [6]. The localization error values, as well as the optimal $\alpha$ values on Imagenet validation set are listed on Tab. 1. It shows our error is slightly higher than Real-time saliency [6] and outperforms all other 5 methods. Note that Real-time saliency[6] utilizes a U-Net [32] architecture which contains encoder and decoder network as mask, and parameters of the masking model is trained over a dataset. The large model size and using a whole dataset for training enables it to achieve relatively higher localization performance.

*4.2.2 Localization Results under Different $\alpha$.* The localization errors of our method over different $\alpha$ value are reported on Fig. 6. The lowest localization error is achieved when $\alpha$ equals 1.5. When $\alpha$ is zero, *i.e.*, we don't apply any post-processing to the generated saliency maps, it illustrates that our method has already achieved 49.9% accuracy, which demonstrates that our interpretation method is able to efficiently suppress the background.

*4.2.3 Localization Results Using Different Layers $l_1$.* In this subsection, we test the object localization performance of interpreting VGG-19 [37] using different layer $l_1$ (see Eq. (6)). $l_1$ could be selected from different higher convolutional layers of CNN, and it is an important factor influencing the performance of the proposed interpretation method. In general, as the convolutional layer of CNN goes deeper, the feature representation at that layer becomes more class discriminative, and thus the likelihood becomes larger to correspond to meaningful objects when mapped to the original image. Specially, we select four higher layers of VGG-19, including *pool4*, *conv5_3*, *conv5_4*, and *pool5*, to test their performance. All the first three layers have 512 channels with shape of (14 × 14), while the last layer have 512 channels with size of (7 × 7). Using the same experimental setting as in Sec. 4.1.1, the optimal $\alpha$ value and its corresponding localization error are obtained at the ImageNet validation set, which are reported in Tab. 2. Among these layers, it's not surprising that *conv5_4* and *pool5* achieves better localization performance than the other two layers, due to they locate at higher layers of CNN. Comparing to *conv5_4*, *pool5* is less accurate, mainly due to the reason that, although the max pooling layer introduces invariance to the neural network, it also leads to the reduced objects localization performance.

|  | *pool4* | *conv5_3* | *conv5_4* | *pool5* |
|---|---|---|---|---|
| $\alpha$ | 0.5 | 1.0 | 1.5 | 1.0 |
| Error(%) | 54.0 | 49.2 | 38.2 | 43.7 |

Table 2: Localization error value, and the optimal $\alpha$ value using different layer of VGG-19 [37].

|  | Center | Grad | Grad-CAM | Ours |
|---|---|---|---|---|
| Accuracy (%) | 0.483 | 0.531 | 0.550 | 0.547 |

Table 3: Pointing game accuracy on PASCAL VOC07 [9].

## 4.3 Pointing Game

In this section, we evaluate the class discriminability of our method by conducting the *Pointing game* experiment [11, 46]. The maximum point is first extracted from each generated saliency map, then according to whether the maximum point falls in one of the ground truth bounding boxes or not, a hit or a miss is counted. The pointing game localization accuracy for each object category is defined as: Acc = $\frac{\#\text{Hits}}{\#\text{Hits}+\#\text{Misses}}$. This process is repeated for all categories and the results are averaged as the final accuracy. The accuracy is evaluated over PASCAL VOC07 [9] test set, which contains 4952 images with multi-label bounding box ground truth. To obtain the classifier for this multi-label classification task, we replace the last fully connected layer of VGG-19 with a new layer which contains 20 output neurons, and then fine-tune the pre-trained VGG-19 model using the training set of VOC07. Following the standard multi-label classification setting, we utilize the Binary Cross Entropy between the target ground truth vector $\mathbf{l}$ and the Sigmoid soft label vector $\mathbf{y}$ as loss function: $l(\mathbf{l}, \mathbf{y}) = -[\mathbf{l} \cdot \log(\mathbf{y}) + (1 - \mathbf{l}) \cdot \log(1 - \mathbf{y})]$. Adam optimizer [19] is utilized to fine-tune the model, with a learning rate of 0.0001. The batch size is set to 64. Only the parameter of the last layer is tuned, and all the other parameters are left unchanged.

We compare the Pointing game performance with Grad [36], Grad-CAM [34], and a baseline method Center which utilizes the center of the image as the maximum point. To obtain the interpretation result, we employ the same empirical setting as in Sec. 4.1.1, except that the parameter $\delta$ in Eq.(9) is set to 10, which works well on this multi-label dataset. Tab. 3 shows that our approach outperforms the center baseline and Grad, and achieves comparable performance with Grad-CAM. Taking into account that most real-life images contain more than one dominative class object in the foreground, the high class-discriminability of the proposed interpretation algorithm is thus an advantage to gain user trust.

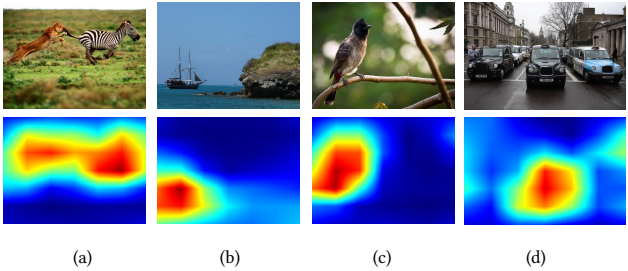

Figure 7: Saliency maps produced by the proposed guided feature inversion method.

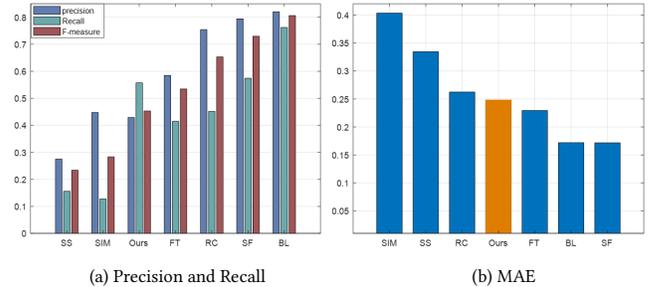

(a) Precision and Recall  (b) MAE

Figure 8: Statistical comparison results of our method as well as 6 state-of-the-art salient object detection methods over MSRA-B dataset.

### 4.4 Evaluation of Guided Feature Inversion

In this section, we evaluate the performance of the first stage of the proposed method, *i.e.*, guided feature inversion, through visualization and by applying it to salient object detection tasks.

*4.4.1 Qualitative Visualization.* Fig. 7 presents the saliency maps generated utilizing guided feature inversion for several inputs. For the first input, our algorithm extracts the locations of hound and zebra, which are all crucial for final prediction. The saliency map of the fourth example most highlights only the centre car, which suffices for the DNN to make classification, although the input contains multiple cars. The result also matches the final prediction very well; the top 10 predictions among 1000 classes are relevant to car. These visualization results reveal that the salient foreground object information is preserved, while the background information which is irrelevant for the final classification task is suppressed. Note that guided feature inversion only utilizes the first half information of the DNN. It is not supposed to produce targeted localization and thus provide explanation for end users. However, it indeed gives us a debugging tool to have some insight into what information is regarded as crucial by the DNN, by visualizing the saliency maps yielded from this stage.

*4.4.2 Weakly Supervised Salient Object Detection.* We further evaluate the proposed guided feature inversion by applying it to *salient object detection*, which targets to detect the full extend of the foregrounds neglecting their categories.

Two metrics are exploited to assess the performance of salient object detection. In the first metric, the generated saliency maps are first binarized using image-dependent threshold. For each saliency map, the threshold is calculated as twice the mean value over all the pixels. The segmented saliency maps are then compared with the ground truth saliency maps using precision, recall and F-measure. Precision is measured as the percentage of the correctly detected pixels to all detected pixels, while the Recall measures the percentage of pixels correctly detected as salient to the ground truth mask. F-measure is also calculated to balance the Precision and Recall, which is defined as following: $F_\beta = \frac{(1+\beta^2) \cdot \text{Precision} \cdot \text{Recall}}{\beta^2 \cdot \text{Precision} + \text{Recall}}$, where the value of $\beta^2$ is set to 0.3 to emphasize the precision [1, 3]. The second metric used to evaluate the saliency detection performance is MAE (mean absolute error), which is defined as the pixel-wise difference between the generated saliency maps and the ground truth masks: $\text{MAE} = \frac{1}{n}\sum_{i=1}^{n}|s_i - L_i|$, where $L$ is the ground truth mask, $n$ is the number of pixels.

We evaluate the salient object detection performance over MSRA-B [26], which is a public benchmark salient object detection dataset containing 5000 images. Note that we neither fine-tune any parameters on this dataset nor provide any post-processing to the generated saliency maps, so as to ensure fair comparison. Our method is compared with six state-of-the-art salient object detection methods, including Signature saliency (SS) [18], Simulation (SIM) [29], Region-based contrast (RC) [3], Frequency-tuned (FT) [1], Saliency filters (SF) [30], Bootstrap learning (BL) [41].

The precision/recall/F-measure, and MAE result are presented in Fig. 8 (a) and (b) respectively. It shows that our method provides slightly better saliency detection result than SS [18], SIM [29] in terms of both two metrics. Although without pixel-wise binary masks as supervision, the proposed guided feature inversion model still generates accurate salient object localization, and achieves comparable performance with state-of-the-art saliency detection approaches. It thus demonstrates that the guided feature inversion part of the proposed method is effective in salient object detection and background removal.

## 5 CONCLUSION

In this work, we propose a class-discriminative DNN interpretation model to explain why a DNN classifier makes a specific prediction for an instance. We show that the inner representations of DNNs provide a tool to interpret and diagnose the working mechanism of individual predictions. By evaluating on ImageNet and PASCAL VOC07 dataset, we demonstrate the interpretability of the proposed model for a variety of CNN models with distinct architectures. The experimental results also validate that the proposed guided feature inversion method performs surprisingly well in preserving the information of all crucial foreground objects, regardless of their category. This untargeted localization has been further applied to general salient object detection task and leaves an interesting direction for future exploration.

## ACKNOWLEDGMENTS

The authors thank the anonymous reviewers for their helpful comments. The work is in part supported by NSF grants IIS-1657196, IIS-1718840 and DARPA grant N66001-17-2-4031. The views and conclusions contained in this paper are those of the authors and should not be interpreted as representing any funding agencies.